\documentclass{Interspeech2025}

 \interspeechcameraready

\title{Employing self-supervised learning models for cross-linguistic child speech maturity classification}

\author[affiliation={1}]{Theo}{Zhang}
\author[affiliation={2}]{Madurya}{Suresh}
\author[affiliation={4}]{Anne S.}{Warlaumont}
\author[affiliation={5}]{Kasia}{Hitczenko}
\author[affiliation={6}]{Alejandrina}{Cristia}
\author[affiliation={2,3}]{Margaret}{Cychosz}

\affiliation{}
{Dept. of Computer Science, UCLA, USA
$^2$Dept. of Linguistics, UCLA, USA
$^3$Dept. of Linguistics, Stanford University, USA
$^4$Dept. of Communication, UCLA, USA
$^5$Department of Computer Science George Washington University, USA
$^6$Département d’études cognitives, ENS, France}

\email{mcychosz@stanford.edu}


\usepackage{comment}
\usepackage{multirow}
\begin{document}

\maketitle
\begin{abstract}
    
    Speech technology systems struggle with many downstream tasks for child speech due to small training corpora and the difficulties that child speech pose. We apply a novel dataset, SpeechMaturity, to state-of-the-art transformer models to address a fundamental classification task: identifying child vocalizations. Unlike previous corpora, our dataset captures maximally ecologically-valid child vocalizations across an unprecedented sample, comprising children acquiring 25+ languages in the U.S., Bolivia, Vanuatu, Papua New Guinea, Solomon Islands, and France. The dataset contains 242,004 labeled vocalizations, magnitudes larger than previous work. Models were trained to distinguish between cry, laughter, mature (consonant+vowel), and immature speech (just consonant or vowel). Models trained on the dataset outperform state-of-the-art models trained on previous datasets, achieved classification accuracy comparable to humans, and were robust across rural and urban settings.
\end{abstract}

\section{Introduction and related work}

In the first years of life, children’s speech becomes increasingly adult-like. 
By about 6-7 months of age, infants start producing sounds that contain both consonant and vowel elements, forming what are known as canonical syllables \cite{ollerEmergenceSpeechCapacity2000}. Canonical syllables continue to make up an increasing proportion of children's vocal productions over the subsequent years, tracking advances in speech maturity. Technology to detect this child speech maturity holds great clinical and educational promise, including the potential ability to identify children at-risk of delay/disorder years before current behavioral techniques permit \cite{ngAutomaticDetectionSpeech2022}. Child speech does not, however, develop the same in all languages. Phonological structures vary dramatically across the world's languages \cite{maddiesonPatternsSounds1984}, and children's early speech reflects this diversity \cite{deboysson-bardiesAdaptationLanguageEvidence1991}. Research in this area has traditionally been constrained by somewhat narrow, linguistically unrepresentative datasets that do not capture the full spectrum of child speech across different linguistic and acoustic environments. 

The bottleneck to progress has been the lack of largescale, carefully-annotated child speech datasets. However, weakly- or self-supervised learning (SSL) models have begun to overcome these size limitations, performing various child speech classification tasks \cite{fanBetterDomainAdaptation2022,gorinSelfsupervisedLearningInfant2023,liEnhancingChildVocalization2024,alfutaisiVCMNetWeaklySupervised2019,zhangAutomatedClassificationChildren2018}. SSL models function by first pre-training on large amounts of unannotated audio or images and then fine-tuning on a smaller amount of annotated data. Here we attempt child speech maturity classification at the utterance level, distinguishing between linguistic (speech-like) versus non-linguistic (cry or laughter), and between mature/canonical (containing a consonant-vowel transition) versus immature/non-canonical (containing just a consonant or vowel) vocalizations \cite{zhangAutomatedClassificationChildren2018,schullerINTERSPEECH2019Computational2019}. A reliable classifier at the individual vocalization level would allow automatic assessment of a child's speech development stage as well as facilitating other analyses such as determining the rates of adult responding to different child vocalization types. 

Previous studies have attempted child vocalization classification, but many focused on basic tasks such as cry detection \cite{michelettiValidatingModelDetect2022,liRobustFamilyInfantAudio2023} and struggled with naturalistic datasets with low accuracy due to limitations and uniformity of the training data. Existing research has been fundamentally limited by datasets that are geographically, linguistically, and acoustically homogeneous, providing only a restricted view of early childhood communication. 
Little work has examined how SSL models classify child speech collected from diverse languages, or in the more realistic contexts in which children actually learn language (c.f. \cite{liEnhancingChildVocalization2024}). Prior work testing these models is largely limited due to the lack of diverse age, language, and recording environment data. 
This is a critical gap because languages differ in the structure of their sound inventories \cite{maddiesonPatternsSounds1984}, and there is significant cross-cultural variation in the acoustic environments in which children develop language---some children spend the majority of their time outdoors meaning that children are systematically exposed to differing amounts of acoustic properties that compete with the speech signal such as wind interference. 

Our approach fundamentally transforms this landscape by introducing a dataset that captures child vocalizations from 25+ languages, spanning radically different acoustic ecologies---from dense, urban centers to remote communities. Not only do we create a more realistic and diverse testing dataset from the novel corpora, we also systematically examine model performance across (e.g. U.S., France) versus rural (e.g. Vanuatu, Bolivia) settings.  
We build on the Interspeech 2019 Computational Paralinguistics Challenge \cite{schullerINTERSPEECH2019Computational2019}, which achieved moderate success (baseline Unweighted Average Recall = 58.7\%).


\section{Child speech corpora}

\subsection{Corpora construction}

We employ two different training corpora: BabbleCorpus (N=11,304 labeled vocalizations from 6 languages)---the training dataset for the current baseline models---and the significantly-expanded SpeechMaturity (N=64,636 vocalizations from 25+ languages) which has never been employed for this task before. BabbleCorpus contains vocalizations from 46 typically-developing children (2-36 months) exposed to a range of mostly genetically-unrelated languages (English, Spanish, Tsimane', Tseltal, Yélî Dnye, and Quechua \cite{casillasCasillasHomeBankCorpus2017,cychoszCychoszHomeBankCorpus2018,bergelsonBergelsonSeedlingsHomeBank2017,cristiaLongformChildcenteredRecordings2018,warlaumontWarlaumontHomeBankCorpus2016,scaffDaylongRecordingsYoung2018,vandamHomeBankOnlineRepository2016}). SpeechMaturity contains vocalizations from 222 children (aged 3-72 months; 90\% typically-developing) exposed to 25+ languages, such as French, Ninde, and Simbo, from 6 different regions. Note that the entire SpeechMaturity corpus contains N=242,004 labeled vocalizations but the ``laughing'' category is under-represented. To mitigate model performance issues due to class imbalance, we thus down-sampled all classes (explained below) except ``laughing,'' to approximately 3x the total ``laughing'' clips (see Table \ref{counts} for counts). 

Data for both corpora were collected using small audio recording devices which the child wore over the course of an entire day (6-16 continuous hours). Vocalizations were extracted by performing voice type diarization upon each recording using either the Language ENvironment Analysis (LENA) system \cite{xuReliabilityLENALanguage2009} or Voice Type Classifier \cite{lavechinOpensourceVoiceType2021}. N=100 (BabbleCorpus) or N=300 (SpeechMaturity) vocalizations/child were sampled from each recording except for \cite{casillasCasillasHomeBankCorpus2017} where all infant vocalizations were hand-segmented. Each vocalization was divided into smaller clips (modal length=500 ms), to remove identifying information, and posted to a public citizen science crowdsourcing website for human annotation. After brief training, citizen scientists listened to the audio clips and classified each as ``crying,'' ``laughing,'', ``canonical,'' ``non-canonical,'' or ``other/junk'' (e.g. no sound, animal sounds, etc.). See \cite{cychoszVocalDevelopmentLargescale2021,hitczenkoDevelopmentCanonicalProportion2023} for further detail. Each vocalization clip was annotated by at least 3 distinct annotators on the platform. For BabbleCorpus, the original corpus creators only kept vocalizations where a \textit{strong} majority of human annotators ($\geq$ 66\%) agreed on the label; the rest of the vocalizations were discarded. For SpeechMaturity, however, we generated two sub-datasets: (1) SpeechMaturity-Cleaned (53,089 clips), which included only vocalizations where likewise a strong majority ($\geq$ 66\%) of human annotators had agreed on the label and (2) SpeechMaturity-Uncleaned (62,000 clips) which included \textit{all} clips from SpeechMaturity-Cleaned, as well as those that were unreliable and/or too difficult to classify and only resulted in the \textit{highest number} of annotator agreements, not necessarily $>$51\% of annotator agreement (e.g. for a clip with 6 annotations, 3 of which were `canonical,' 2 were `laughing', and 1 was `crying' would be classified as `canonical'). This is a critical distinction between BabbleCorpus---the current state-of-the-art dataset for this task---and SpeechMaturity: BabbleCorpus represents an idealized set of training data that is not representative of the high variability in children's vocalizations because a portion of the data were removed in post-processing. SpeechMaturity-Cleaned represents a similarly idealized set of test data but across a more diverse spread of ages, languages, and recording environments, while SpeechMaturity-Uncleaned represents the truest test of how the proposed models would extrapolate to new samples. We compare the effect of this data removal step by training the proposed models upon the idealized SpeechMaturity-Cleaned but evaluating model performance upon both SpeechMaturity-Cleaned and -Uncleaned.

\subsection{Corpora pre-processing}

For precise comparison of our approaches with previous models that were trained using BabbleCorpus, we replicate the train/dev/test split from the original 2019 Paralinguistics Challenge for BabbleCorpus (Table \ref{counts}; train/dev/test=35/32/33, with child-disjunct folds). We apply a more traditional 80/10/10 for SpeechMaturity, again with child-disjunct folds, and ensuring that all child ages and languages were approximately evenly represented across folds. 
Audio clips were converted to mono audio arrays, resampled to 16kHz, and 0-padded around the center such that all arrays contained 9217 elements, which was the maximum length after mono conversion and resampling. This ensured all model inputs were uniform, and no data were lost through truncation. Scripts to replicate our models are available at \url{github.com/spoglab-stanford/w2v2-pro-sm/tree/main/speechbrain/recipes/W2V2-LL4300-Pro-SM}.

{\large
\begin{table}[htbp]
\caption{Class distribution for child speech corpora employed in model training}
\vspace{-1 em}
\label{counts}
\begin{center}
\resizebox{\columnwidth}{!}{%
\begin{tabular}{|l|ccc|ccc|c|}
\hline
\multirow{2}{*}{\textbf{Class}} & \multicolumn{3}{c|}
{\textbf{BC$^{\mathrm{a}}$}} & \multicolumn{3}{c|}
{\textbf{\shortstack{SM-C$^{\mathrm{b}}$}}}  & \multicolumn{1}{c|}{\textbf{\shortstack{SM-U$^{\mathrm{c}}$}}} \\
\cline{2-8}
& Train & Dev & Test & Train & Dev & Test & Test \\
\hline
Crying & 243 & 163 & 263 & 9830 & 1177 & 1098 & 1116 \\
Laughing & 46 & 41 & 62 & 3491 & 388 & 356 & 358\\
Canonical & 444 & 378 & 604 & 9762 & 1262 & 1226 & 1674\\
Non-Canonical& 1437 & 1678 & 1370 & 9766 & 1232 & 1252 & 1867\\
Junk & 1826 & 1357 & 1392 & 9838 & 1226 & 1185 & 1185\\
\hline
\textbf{Total} & 3996 & 3617 & 3691 & 42687 & 5285 & 5117 & 6200\\
\hline
\end{tabular}
}
\label{counts}
\end{center}
\begin{flushleft}
\footnotesize
\vspace{-1 em}
$^{\mathrm{a}}$\textbf{BC}: BabbleCorpus - Reflects original train/dev/test splits from previous attempts. \\
$^{\mathrm{b}}$\textbf{SM-C}: SpeechMaturity-Cleaned - Reflects data after down-sampling to mitigate class imbalance. \\
$^{\mathrm{c}}$\textbf{SM-UC}: SpeechMaturity-Uncleaned - Only the test set from SpeechMaturity-Uncleaned was used in experiments. Also reflects data after down-sampling to mitigate class imbalance.\\
\end{flushleft}
\vspace{-2 em}
\end{table}
}

\section{Model architectures}

We employed three different Wav2Vec2 models of varying size and complexity for the task of child speech maturity classification: {\textit{\textbf{W2V2-base}}}, {\textit{\textbf{W2V2-LL4300h}}}, and {\textit{\textbf{W2V2-LL4300-Pro}}}. The three models were pretrained in different ways and fine-tuned on either the BabbleCorpus or SpeechMaturity-Cleaned datasets (explained below). \textit{\textbf{W2V2-base}} was pre-trained on thousands of hours of unlabeled LibriSpeech (English) \cite{panayotovLibrispeechASRCorpus} with 12 transformer layers, hidden dimension of 768, inner dimension of 3,072, and 8 attention heads \cite{baevskiWav2vecFrameworkSelfSupervised2020}). \textit{\textbf{W2V2-base}} passes outputs from a CNN feature extractor through a transformer architecture to develop contextualized speech representations (see \cite{baevskiWav2vecFrameworkSelfSupervised2020} for further description). The second model was \textit{\textbf{W2V2-LL4300h}}, which pre-trained \textit{\textbf{W2V2-base}} on 4300h of daylong home-based audio recordings of children under 5 years old acquiring English \cite{liRobustFamilyInfantAudio2023}. Finally, \textit{\textbf{W2V2-LL4300h}} incorporates the auxiliary task of child speech phoneme recognition within \textit{\textbf{W2V2-LL4300h}} \cite{liEnhancingChildVocalization2024}. To achieve this, the hidden features from a model that generates phonetic pseudo-reference transcripts are fused with \textit{\textbf{W2V2-LL4300h}} by adding a linear layer to a middle transformer layer of \textit{\textbf{W2V2-LL4300h}}. It is then trained using CTC to generate hypothesis transcripts that match the pseudo-reference transcripts with minimum cross-entropy. Essentially, the auxiliary task requires \textit{\textbf{W2V2-LL4300h}} to simultaneously learn and incorporate children's phonetic representations along with predicting the target 5 labels, a change that results in the \textit{\textbf{W2V2-LL4300-Pro}} model. \cite{liEnhancingChildVocalization2024}. \textit{\textbf{W2V2-LL4300-Pro}} is more complex compared to \textit{\textbf{W2V2-LL4300h}} due to these additions, and thus is the most complex model upon which we conduct experiments.

All three models were then subsequently fine-tuned on either the BabbleCorpus dataset or the SpeechMaturity-Cleaned dataset (noted in Table \ref{uar_past_models} with either -BC or -SM). 
For \textit{\textbf{W2V2-base}}, audio data were inputted into the model's feature extractor; those features were then inputted into the pre-trained model's transformer architecture for classification in train/dev/test batches as outlined in Table \ref{counts}. Data were processed in batches of 32, via random sampling during training. For \textit{\textbf{W2V2-LL4300h}} and \textit{\textbf{W2V2-LL4300-Pro}}, we replicated the training and testing environments with the provided pre-trained checkpoints from \cite{liEnhancingChildVocalization2024}. Training for all three models was conducted synchronously on 10 CPU cores for 10 epochs (learning rate=3e-5 for \textit{\textbf{W2V2-base}} and 1e-5 for \textit{\textbf{W2V2-LL4300h}} and \textit{\textbf{W2V2-LL4300-Pro}}). The best-performing epoch for each model was then used for testing.

\section{Results}

Following previous classification models for this task (e.g. \cite{yehUsingAttentionNetworks2019}), model performance was evaluated using the unweighted average recall (UAR), a metric that takes the mean of recall values for each class, giving equal importance to each class regardless of its size. It is thus well-suited for multi-class classification tasks, especially when classes are imbalanced as they are here \cite{keesingAcousticFeaturesNeural2021} (Table \ref{uar_with_explanation_underneath}). 
Results highlight two key contributions of this work, one regarding data and the other model architecture. Across models and test sets, performance increases when fine-tuning on the SpeechMaturity dataset compared to BabbleCorpus, with increases that go well beyond simply matching training and test sets. For example, each model type gained between 7 and 30\% when the fine-tuning set was Speech-Maturity-Cleaned and the test sets were either of the other two. As for architecture, we replicate the observations that pretraining on relevant data and incorporating a phonetic task improves performance when compared with W2V2-base \cite{liEnhancingChildVocalization2024}, but further show that the relative gain depends crucially on the fine-tuning dataset (e.g., max. 26.4\% for BabbleCorpus versus max. 4.6\% for SpeechMaturity, both when testing on BabbleCorpus). Finally, our conditions significantly surpassed previous state-of-the-art performance (best UAR for \textit{\textbf{W2V2-LL4300-Pro-SM}} at 74.2\%), even when holding the test set constant (same model, 68.6\% versus 54-64.3\%; see Table 2).

\begin{table}
\caption{State-of-the-art model performance as reported in all prior publications attempting this classification task, all tested on the BabbleCorpus dataset (the only extant at the time). The best Test performance corresponds to Li et al.}
\begin{center}
\vspace{-1 em}
\resizebox{\columnwidth}{!}{%
\begin{tabular}{|c|c|c|c|}
\hline
\textbf{Model}&\multicolumn{3}{|c|}{\textbf{UAR (\%)}} \\
\cline{2-4} 
& \textbf{\textit{Train}}& \textbf{\textit{Dev}}& \textbf{\textit{Test}} \\
\hline
\multicolumn{4}{|c|}{\textbf{Dataset: BabbleCorpus}} \\
\hline
Challenge Baseline$^{\mathrm{a}}$ \cite{schullerINTERSPEECH2019Computational2019} & * & 54.0 & 58.7 \\
\hline
Yeh et al.$^{\mathrm{a}}$ \cite{yehUsingAttentionNetworks2019} & * & 61.3 & 62.4 \\
\hline
Gosztolya$^{\mathrm{a}}$ \cite{gosztolyaUsingFisherVector2019}& * & 58.7 & 59.5 \\
\hline
Kaya et al.$^{\mathrm{a}}$ \cite{kayaCombiningClusteringFunctionals2020} & * & 60.1 & 61.4 \\
\hline
Li et al. \cite{liEnhancingChildVocalization2024} & * & \textbf{70.4} & \textbf{64.6} \\
\hline
\multicolumn{4}{l}{$^{\mathrm{a}}$Performance on Train set not reported in original paper.} \\
\end{tabular}
}
\vspace{-2 em}
\label{uar_past_models}
\end{center}
\end{table}

\begin{table}[h!]
\caption{UAR (\%) across fine-tuning (rows) and testing sets (columns) for our best-performing models. Performance metrics listed in the BC (BabbleCorpus) column correspond to the same Test data as in the Test column in Table 2.}
\begin{center}
\vspace{-1 em}
\resizebox{\columnwidth}{!}{%
\begin{tabular}{|c|c|c|c|}
\hline
\textbf{Model} & \multicolumn{3}{|c|}{\textbf{UAR (\%)}} \\
\cline{2-4} 
 & \textbf{BC}$^{\mathrm{a}}$ & \textbf{SM-C}$^{\mathrm{b}}$ & \textbf{SM-UC}$^{\mathrm{c}}$ \\
\hline
\multicolumn{4}{|c|}{\textbf{Fine-tuning Dataset: BabbleCorpus}} \\
\hline
W2V2-base-BC & 34.4 & 29.9 & 29.4 \\
\hline
W2V2-LL4300h-BC & 60.2 & 50.0 & 48.1 \\
\hline
W2V2-LL4300-Pro-BC & 60.8 & 51.7 & 49.8 \\
\hline
\multicolumn{4}{|c|}{\textbf{Fine-tuning Dataset: SpeechMaturity-Cleaned}} \\
\hline
W2V2-base-SM & 64.2 & 71.0 & 68.4 \\
\hline
W2V2-LL4300h-SM & 66.6 & 73.8 & 71.5 \\
\hline
W2V2-LL4300-Pro-SM & \textbf{68.6} & \textbf{74.2} & \textbf{71.9} \\
\hline
\end{tabular}
}
\end{center}
\label{uar_with_explanation_underneath}
\begin{flushleft}
\vspace{-2 em}
$^{\mathrm{a}}$ \textbf{BC}: BabbleCorpus testing set. \\
$^{\mathrm{b}}$ \textbf{SM-C}: SpeechMaturity-Cleaned testing set. \\
$^{\mathrm{c}}$ \textbf{SM-UC}: SpeechMaturity-Uncleaned testing set. \\
\end{flushleft}
\vspace{-1 em}
\end{table}

\begin{figure}[!htpb]
\centerline{\includegraphics[width=.85\linewidth]{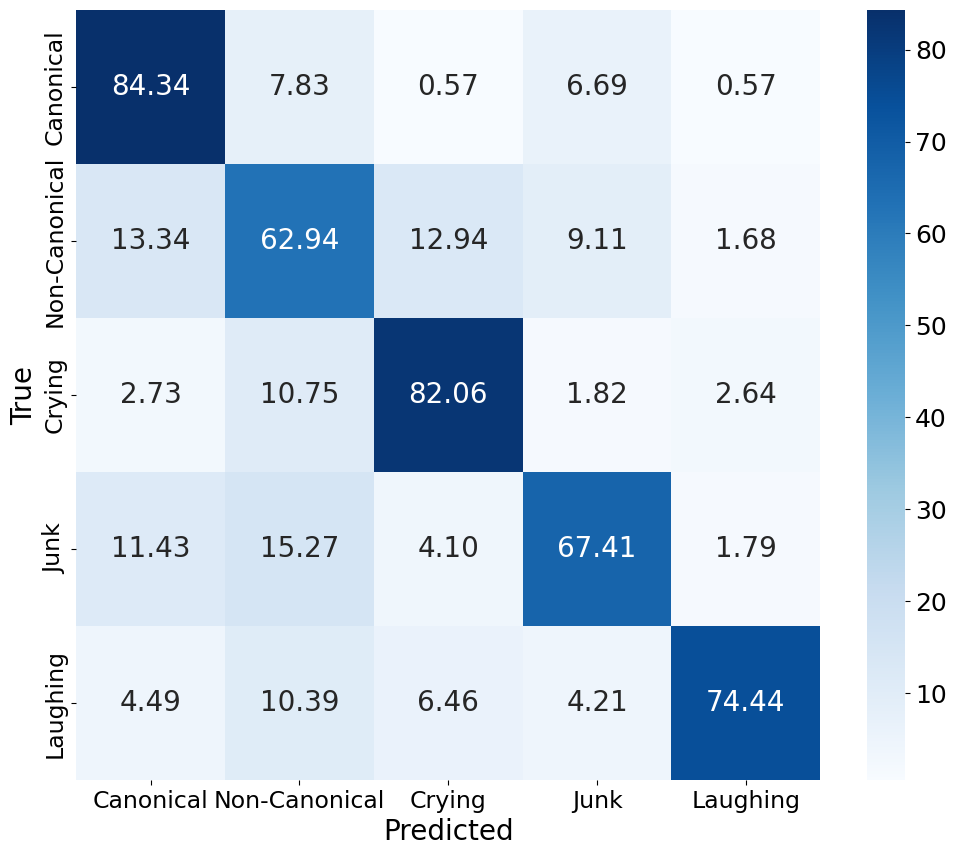}}
\caption{Confusion matrix of the SpeechMaturity-Cleaned test set predictions for \textit{\textbf{W2V2-LL4300-Pro}} finetuned on SpeechMaturity-Cleaned.}
\label{transformer}
\vspace{-1 em}
\end{figure}
For by-category results, Figure \ref{transformer} shows the final confusion matrix over the SpeechMaturity-Cleaned test set for \textit{\textbf{W2V2-LL4300-Pro-SM}} which was fine-tuned on SpeechMaturity-Cleaned. The results of \textit{\textbf{W2V2-LL4300-Pro-SM}} far surpass the state-of-the-art (trained on the smaller, less representative BabbleCorpus) which achieved a UAR=67.7\% for the speech category ``canonical'' and UAR=42.5\% for ``non-canonical'' in \cite{yehUsingAttentionNetworks2019}, while \textit{\textbf{W2V2-LL4300-Pro-SM}} achieved a UAR=84.3\% on ``canonical'' and UAR=62.9 on ``non-canonical.''

\subsection{Classification accuracy of models versus human annotators}

To set a benchmark for model performance and rigorously assess the performance of our model in a way that previous work has not, we conducted an agreement analysis on \textit{\textbf{W2V2-LL4300-Pro-SM}}. We assessed inter‐human annotator agreement using a weighted Fleiss' kappa metric for the SpeechMaturity-Cleaned and -Uncleaned datasets. 
We employed a custom weighting scheme to reflect the relative importance of different category distinctions in the classification task \cite{fleissMeasuringNominalScale1971}, assigning higher weights to the canonical versus non-canonical distinction (the most important linguistic classification distinction in this task), as well as between speech-like categories (canonical, non-canonical) versus non speech-like vocalizations (cry, laugh, and junk). 
Using this weighted scheme, we found a weighted Fleiss' kappa of K = 0.375  (95\% CI: 0.361-0.388) for SpeechMaturity-Cleaned and K = 0.271 (0.259-0.284) for SpeechMaturity-Uncleaned for agreement between 5 or fewer annotators. Unsurprisingly, agreement improved when comparing between vocalizations annotated by 3 or fewer annotators for SpeechMaturity-Uncleaned (K = 0.282, CI: 0.106-0.458) and -Cleaned (K = 0.457, CI: 0.402, 0.512). These results indicate fair to moderate levels of agreement among human annotators---which is unsurprising for a task of this complexity---and suggest a benchmark by which to compare model performance.

\begin{figure}
\centerline{\includegraphics[width=.7\linewidth]{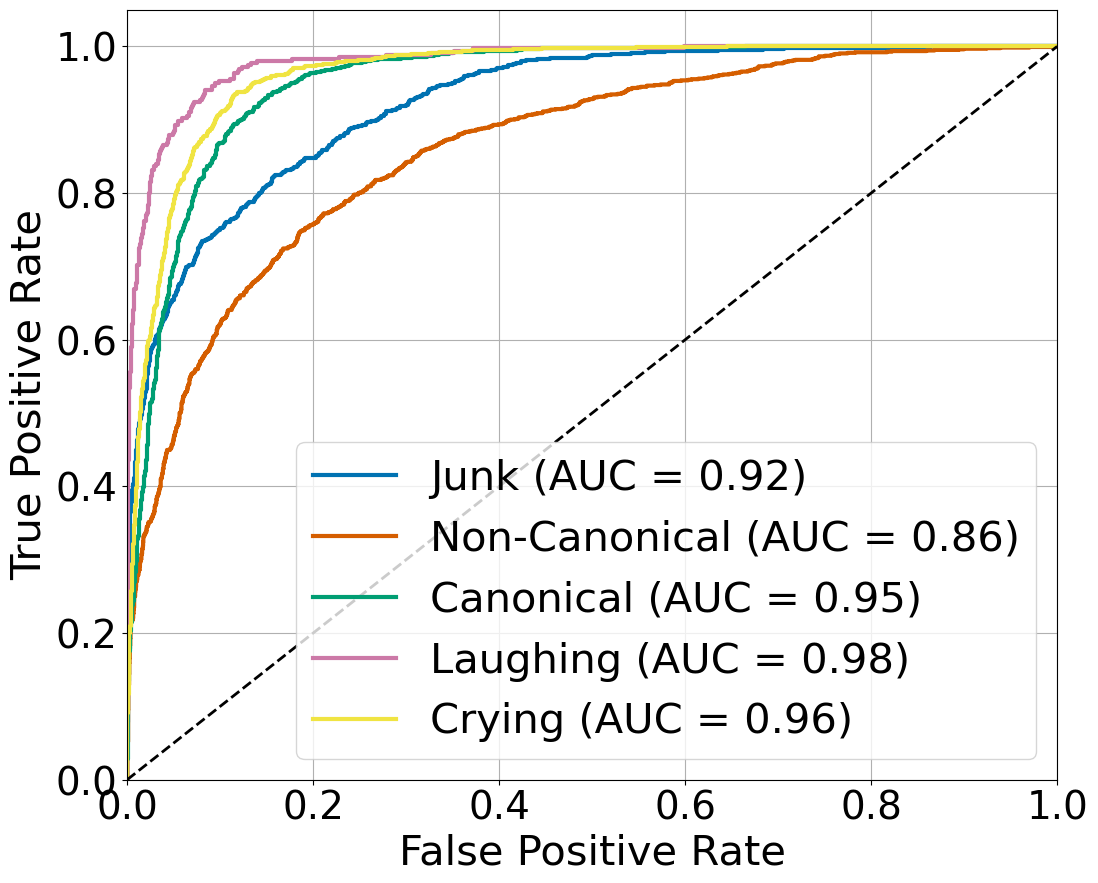}}
\centerline{\includegraphics[width=.7\linewidth]{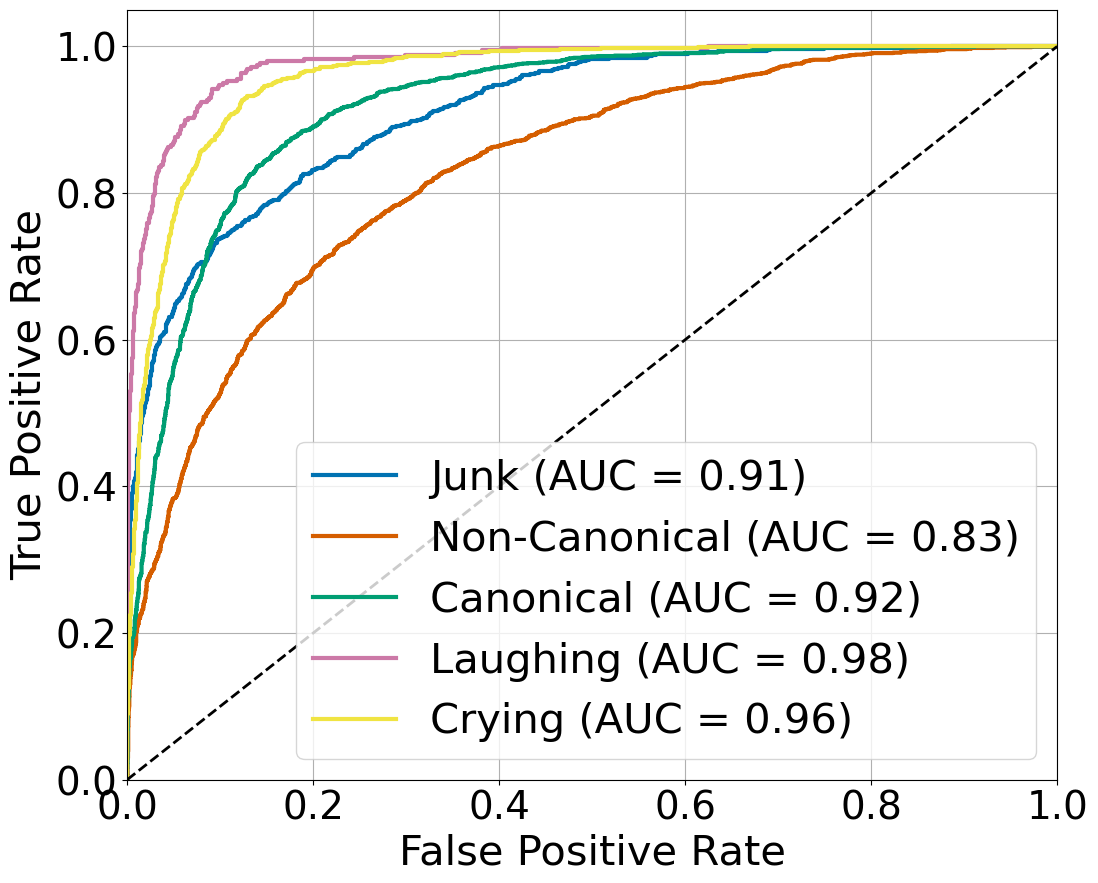}}
  \caption{ROC curves achieved by \textit{\textbf{W2V2-LL4300-Pro-SM on SpeechMaturity-Cleaned}} (top) and \textit{\textbf{W2V2-LL4300-Pro-SM on SpeechMaturity-Uncleaned}} (bottom) for classification of canonical (green), non-canonical (red), crying (yellow), laughing (pink), and junk (blue) in comparison to human annotators.}
  \vspace{-1 em}
  \label{ROC}
\end{figure}

Next, we compared human annotator performance to \textit{\textbf{W2V2-LL4300-Pro-SM}} by calculating weighted Cohen's kappa between \textit{\textbf{W2V2-LL4300-Pro-SM}} and each annotator of each vocalization over SpeechMaturity-Cleaned and -Uncleaned. The average weighted Cohen's kappa was K = 0.478 (SD = 0.012) for SpeechMaturity-Cleaned and K = 0.406 (SD = 0.015) for SpeechMaturity-Uncleaned, indicating moderate levels of agreement between \textit{\textbf{W2V2-LL4300-Pro-SM}} and human annotators. The standard deviation reflects the variability in agreement between the model and different annotators. Most critically, these levels of agreement approach and/or surpass the level of multiple human inter-rater agreement computed above. Figure \ref{ROC} displays the ROC curves and respective by-category AUC scores. For the model tested on SpeechMaturity-Cleaned, the AUC scores indicate excellent to outstanding discrimination across all categories with particularly strong classification performance for canonical vocalizations. For SpeechMaturity-Uncleaned we likewise found excellent to outstanding AUC scores. 

\subsection{Classification accuracy of the SSL models by language learning environment}


\textit{\textbf{W2V2-LL4300-Pro-SM}} tested over SpeechMaturity-Cleaned achieved a UAR of 70.7\% in urban environments and 67.8\% in rural environments, showing a robustness to data from the vastly different language and acoustic learning environments of children from around the world (Table \ref{urbanrural}). The discrepancy in UARs by environment could be due to a number of reasons including wind interference (due to increased time spent out of doors) or more multi-party and/or overlapping speech, both of which have been documented during psycholinguistic and ethnographic fieldwork conducted in the language learning environments of children in the rural areas of this corpus (e.g. \cite{scaffCharacterizationChildrenVerbal}). Further exploration of the causes of the discrepancies by rural versus urban settings is an avenue for future research.  

\begin{table}[!htpb]
\caption{Distribution and Performance by Language Environment for SpeechMaturity-Cleaned. Upper= clip counts by split in urban v. rural. Lower= UARs. }
\begin{center}
\vspace{-1 em}
\resizebox{\columnwidth}{!}{%
\begin{tabular}{|c|c|c|c|}
\hline
\textbf{Split/}&\multicolumn{3}{|c|}{\textbf{Language Environment}} \\
\cline{2-4} 
\textbf{Metric} & \textbf{\textit{Urban}}& \textbf{\textit{Rural}}& \textbf{\textit{Total}} \\
\hline
Train Clips & 8508& 32723& 41231\\
\hline
Dev Clips & 759& 4283& 5042\\
\hline
Test Clips & 681& 4436& 5117\\
\hline
Total Clips & 9948 & 41442& 51390\\
\hline
UAR (SD)$^{\mathrm{a}}$ &  \textbf{70.7} ($<$0.01)& \textbf{67.8} ($<$0.01)& -\\
\hline
\multicolumn{4}{l}{$^{\mathrm{a}}$UAR: Unweighted Average Recall (\%); SD: Standard Deviation}
\end{tabular}
}
\vspace{-2 em}
\label{urbanrural}
\end{center}
\end{table}

\section{Discussion}

SpeechMaturity represents a significant shift in child speech research, challenging existing methodological constraints in computational studies of child speech development. By capturing child vocalizations across 25+ languages and dramatically diverse acoustic environments---from urban centers in industrialized communities to remote communities in Vanuatu and Papua New Guinea---this corpus provides an unprecedented window into the global landscape of early vocal development. The dataset is openly available for other researchers to use \cite{hitczenkoSpeechMaturityDatasetunderreview} 
and build tools with, allowing new questions about early communication to be answered in a cross-linguistically diverse sample.

To demonstrate the potential of this dataset to resolve long-standing challenges in child speech technology, we applied three transformer models to child speech maturity classification. When trained on the comprehensive SpeechMaturity dataset (N=64,636 audio clips from 222 children acquiring one or more of 25+ languages), the models outperformed those trained on smaller, less representative datasets (BabbleCorpus). This performance gain reflects SpeechMaturity's ecological richness: by including children from vastly different linguistic and acoustic settings, SpeechMaturity enables more robust generalizable insights into early speech development. Notably, model performance consistently improved after fine-tuning on SpeechMaturity, regardless of model complexity. This underscores the dataset's fundamental value: it captures variability in child speech that previous, smaller and more limited corpora overlooked. Even the lowest-complexity model, \textit{\textbf{W2V2-base-SM}} which required no additional pre-training or auxiliary tasks and took the least amount of memory to fine-tune and test, performed comparably to either of the more computationally-intensive models, suggesting that the dataset's diversity may be more impactful for success at this task than overall model architecture or sophistication. The best performing model, \textit{\textbf{W2V2-LL4300-Pro-SM}} fine-tuned on the novel SpeechMaturity dataset exceeds previously published state-of-the-art solutions and (1) was robust even on a dataset that consisted of ``noisier'' clips (SpeechMaturity-Uncleaned), (2) had acceptable levels of agreement with human annotators and strong AUC values, and (3) had consistently high levels of performance between rural and urban child rearing environments.

\bibliographystyle{IEEEtran}
\bibliography{mylib_v2}
\end{document}